\documentclass{article}

\usepackage[final]{corl_2026} %

\usepackage{amsmath}
\usepackage{amssymb}
\usepackage{booktabs}
\usepackage{multirow}
\usepackage{graphicx}
\usepackage{xcolor}
\usepackage{microtype}
\usepackage{subcaption}
\usepackage{wrapfig}
\usepackage{xspace}
\usepackage{tikz}
\usepackage{makecell}
\usetikzlibrary{arrows.meta, positioning, decorations.pathreplacing, fit, backgrounds, calc}
\usepackage{pgfplots}
\usepackage{pifont}
\pgfplotsset{compat=1.18}

\definecolor{task1}{HTML}{00994D}
\definecolor{task2}{HTML}{66B2FF}
\definecolor{task3}{HTML}{FF3333}

\usepackage{enumitem}
\setlist[itemize]{leftmargin=1.5em, itemsep=2pt, topsep=2pt, parsep=0pt}
\setlist[enumerate]{leftmargin=1.8em, itemsep=2pt, topsep=2pt, parsep=0pt}

\newcommand{\ours}{\textsc{LEGS}\xspace}
\newcommand{\gsplat}{3DGS\xspace}
\newcommand{\groot}{GR00T~N1.6\xspace}
\newcommand{\pihalfpoint}{$\pi_{0.5}$\xspace}

\newcommand{\psizero}{$\psi_0$\xspace}

\newcommand{\cmark}{\textcolor{black!75}{\ding{51}}}
\newcommand{\xmark}{\textcolor{black!45}{\ding{55}}}
\newcommand{\blfootnote}[1]{%
  \begingroup
  \renewcommand\thefootnote{}\footnote{#1}%
  \addtocounter{footnote}{-1}%
  \endgroup
}

\title{%
\Large
LEGS: Fine-Tuning Teleop-Free VLAs for Humanoid Loco-manipulation in an Embodied Gaussian Splatting World
}

\author{%
\makebox[\linewidth][c]{%
\begin{tabular}{@{}cccc@{}}
\textbf{Hojune Kim} & \textbf{Timothy Chen} & \textbf{Jiankai Sun} & \textbf{Lars W. Osterberg}\\
\multicolumn{4}{c}{%
  \begin{tabular}{@{}ccc@{}}
    \textbf{Qianzhong Chen} & \textbf{Ke Wang} & \textbf{Mac Schwager}
  \end{tabular}%
}\\
\end{tabular}}\\[0.8em]
\makebox[\linewidth][c]{Stanford University}\\
}

\begin{document}
\maketitle
\blfootnote{correspondence to \href{mailto:hojune@stanford.edu}{\texttt{hojune@stanford.edu}}}

\vspace{-0.5cm}

\begin{figure}[h]
    \centering
    \vspace{-1cm}
    \includegraphics[clip, trim={0 0cm 0 0cm}, width=\linewidth]{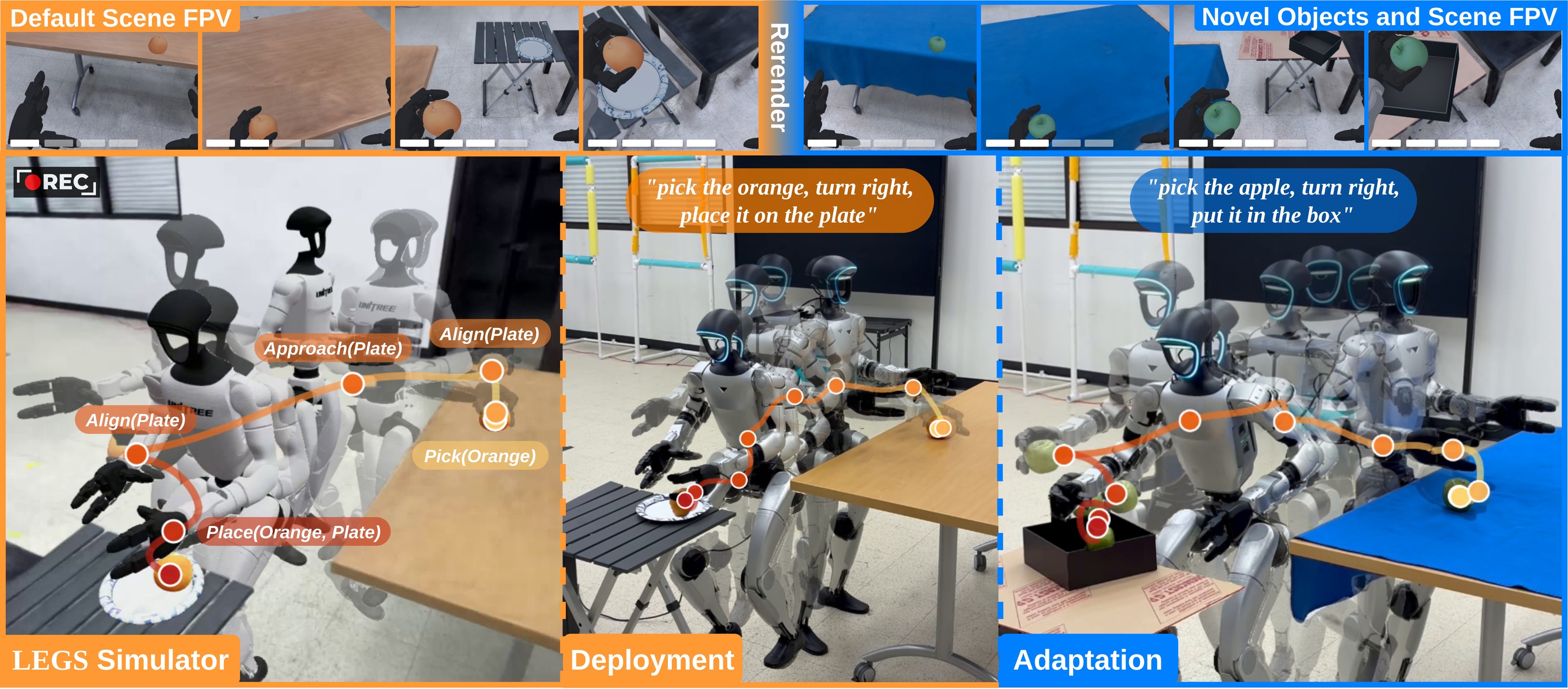}
    \caption{\textbf{L}oco-manipulation via \textbf{E}mbodied \textbf{G}aussian \textbf{S}platting (\ours) combines the physical fidelity of MuJoCo, the visual fidelity of 3D Gaussian Splatting, and single-image object reconstruction from SAM3D to generate inexpensive humanoid training data without any real-robot data collection. Vision-Language-Action (VLA) models fine-tuned on \ours data transfer zero-shot to the real robot and outperform those trained on teleoperation or mesh-only simulation.}
    \label{fig:teaser}
\end{figure}

\begin{abstract}
Training vision-language-action (VLA) policies for humanoid loco-manipulation is constrained by the high cost and complexity of collecting human teleoperation demonstrations. VLA policies fine-tuned in simulators have so far transferred poorly to humanoid loco-manipulation tasks.
We present \ours (\textbf{L}oco-manipulation via \textbf{E}mbodied \textbf{G}aussian \textbf{S}platting), a hybrid simulator that composites a mesh foreground (robot, objects, props) over a photorealistic 3D Gaussian Splatting (\gsplat) background reconstructed from a handheld scene capture.
\ours uses a procedural motion-primitive generator to synthesize labeled demonstrations at scale without human teleoperation, and a two-stage color calibration that aligns renders to the deployment camera.
On a Unitree G1 humanoid robot, across three pick-and-place tasks of increasing whole-body difficulty and three VLA backbones ($\psi_0$, $\pi_{0.5}$, GR00T N$1.6$), a policy trained purely on \ours data matches or exceeds one trained on human teleoperation demos on every experiment. It also outperforms a mesh-only baseline trained on the same motions without the calibrated \gsplat frontend.
Because motion is recorded independently of appearance, the same demonstrations can be re-rendered for a target scene and objects at over 15$\times$ lower cost than teleoperation. When both the objects and the scene change, a policy trained on \textsc{LEGS-aug} data re-rendered for that target maintains task success, while the teleop baseline fails entirely. Code and videos are available at \url{https://hojjunekim.github.io/legs/}.
\end{abstract}


\section{Introduction}
\label{sec:intro}

The humanoid form factor is uniquely suited to operate alongside humans: shared morphology allows direct use of human-built environments, tools, and infrastructure without retrofitting.
Recent vision-language-action (VLA) foundation models \citep{black2024pi0, pi05, grootn12024,firoozi2025foundation} show that large-scale pre-training unlocks impressive zero-shot manipulation, but adapting them to a specific humanoid platform still requires fine-tuning demonstrations covering both manipulation and locomotion skills.
\emph{Teleoperation}~\citep{omnih2o2024,twist2025, homie2025} binds data to operator labor and robot uptime, while \emph{human-demo retargeting}~\citep{umi2024,humanplus2024,wholebodyvla2025} removes the robot from the loop, but still requires per-scene egocentric capture.
Any appearance shift at deployment requires human labor for each new episode, and operator and hardware fatigue make extended collection sessions increasingly costly (Table~\ref{tab:teleop_vs_legs}).

Collecting data in simulation mitigates human labor, but conventional simulators introduce a \emph{visual gap}: mesh-based rendering produces images whose distribution diverges from real camera footage, and even strong mesh-synthetic pipelines built on SAM3D~\citep{sam3d2024} reconstructions suffer from material-shading mismatch, simplified lighting, and aliasing that bias the visual encoder.
3D Gaussian Splatting~(3DGS)~\citep{kerbl3Dgaussians}, in contrast, is trained to minimize rendering error against \emph{real} photographs from the target scene, substantially reducing the visual sim-to-real gap.

We propose \ours, a photorealistic simulator that generates humanoid loco-manipulation training data without any teleoperation. \ours builds on established real-to-sim components (MuJoCo physics, \gsplat, and SAM3D object reconstruction); our contribution is their composition into a teleop-free data pipeline for humanoid VLA fine-tuning, together with an extensive real-robot study of when such data transfers. \ours renders a dynamic mesh foreground (robot, manipulable objects, props) over a static \gsplat background reconstructed from a handheld scene capture, and composites the two through a deterministic two-stage color calibration that aligns the render to the deployment camera. Because the foreground motion is recorded as a command stream independent of the rendering inputs, a procedural generator synthesizes labeled demonstrations at scale, and the same motion dataset re-renders under new backgrounds and object meshes at GPU cost alone. We evaluate \ours on a Unitree~G1 across three pick-and-place tasks of increasing whole-body difficulty and three VLA backbones (\psizero, \pihalfpoint, \groot), benchmarked against human teleoperation and a mesh-only synthetic baseline. Our contributions are as follows:

\begin{itemize}
    \item \textbf{A teleop-free data pipeline for humanoid VLA fine-tuning.}
    \ours generates labeled loco-manipulation data entirely in simulation, with no human teleoperation or per-episode human intervention; task semantics are instead specified once per task as a plan over shared motion primitives (Appendix~\ref{app:supervision}). A policy fine-tuned purely on \ours data matches or exceeds one fine-tuned on human teleoperation across all three tasks and backbones, including a budget-matched 50-demonstration comparison and a long-horizon task on which teleop fails entirely. Against a mesh-only baseline that shares the same motion data but lacks the calibrated \gsplat frontend, \ours wins on every experiment at all dataset sizes.

    \item \textbf{A parallelizable re-rendering workflow for low-labor adaptation to target appearances.}
Because motion is decoupled from appearance, \ours re-renders one recorded motion dataset under swapped backgrounds and object meshes. For each additional appearance condition with an existing scene capture, \ours replaces $>$1.5~hr of human teleop labor (Table~\ref{tab:teleop_vs_legs}) with $\sim$0.1~hr of GPU time on consumer-grade hardware. The resulting \textsc{LEGS-aug} dataset, re-rendered for the target objects and scene, retains task success under that combined shift, while teleop fails.

    \item \textbf{Extensive real-robot evaluation across three SOTA VLAs and four data conditions (1{,}230 trials, plus 240 matched-camera trials).}
    The same findings reproduce across \psizero, \pihalfpoint, and \groot fine-tuned from their public checkpoints, suggesting they are a property of the data pipeline rather than any single VLA. Because \ours produces a general-purpose humanoid loco-manipulation demonstration dataset, the same recorded-and-augmented data can fine-tune other VLA backbones that accept the same command interface.

\end{itemize}

\begin{table}[h]
\centering
\small
\setlength{\tabcolsep}{6pt}
\renewcommand{\arraystretch}{1.15}
\caption{
  \textbf{Data sources for humanoid loco-manipulation policy training.}
  Teleop and \ours costs measured on Task~3 (long-horizon, 50 episodes); new condition cost is the marginal cost of producing 50 episodes for one additional appearance condition with an existing scene capture. \ours costs cover data execution and rendering only; they exclude the one-time framework engineering and the per-task setup (Appendix~\ref{app:supervision}). Human-demo cost is extrapolated from the per-episode timing ratio reported in~\citep{egohumanoid2026} on the Unitree~G1.
}
\label{tab:teleop_vs_legs}
\begin{tabular}{@{}l >{\centering\arraybackslash}p{2.0cm} >{\centering\arraybackslash}p{1.4cm} >{\centering\arraybackslash}p{1.8cm} >{\centering\arraybackslash}p{1.4cm} >{\centering\arraybackslash}p{2.2cm}@{}}
\toprule
 & No on-robot data collection & No per-episode human & Re-renderable to new scenes & Initial cost (hr) & New condition cost (hr) \\
\midrule
Teleop~\citep{omnih2o2024,twist2025, lu2025}        & \xmark & \xmark & \xmark & 1.5 & $>$1.5 \\
Human demo~\citep{wholebodyvla2025,egohumanoid2026,humi2026}    & \cmark & \xmark & \xmark & $\sim$1.0 & $>$1.0 \\
\ours (ours)                                         & \cmark & \cmark & \cmark & \textbf{0.5} & \textbf{0.1} \\
\bottomrule
\end{tabular}
\end{table}

\section{Related Work}
\label{sec:related}

\paragraph{Humanoid loco-manipulation and VLA foundation models.}
Producing the whole-body motion that positions a humanoid for loco-manipulation typically relies on shadowing or wearable-device teleoperation~\citep{omnih2o2024,twist2025} or decoupling locomotion from manipulation~\citep{lu2025}, all of which require on-robot data collection. Robot-free human-demonstration systems~\citep{wholebodyvla2025,egohumanoid2026,humi2026} avoid teleoperation but remain expensive to scale due to per-scene human capture. In parallel, vision-language-action foundation models have advanced rapidly~\citep{pi05,grootn12024,psi0}. We instead generate motion procedurally in simulation and fine-tune three pretrained VLAs on the resulting data.
\vspace{-1em}
\paragraph{Teleoperation-free humanoid loco-manipulation.}
Concurrent work explores humanoid loco-manipulation without on-robot teleoperation through several paradigms: sim-augmented behavior cloning from a single seed demonstration~\citep{demohlm2025}, distillation of a privileged RL teacher into a vision-based student~\citep{viral2025,doorman2025}, and reference-tracking RL from optimization-derived trajectories~\citep{opt2skill2024}. Each removes the operator but still requires per-task specification: a hand-collected seed, a hand-engineered reward, or a hand-designed trajectory. \ours also requires per-task specification, in the form of a task plan over shared motion primitives (Appendix~\ref{app:supervision}), but involves no human teleoperation or per-episode human intervention---no seed demonstration, no human video, no per-task reward---and fine-tunes pretrained VLA backbones rather than training from scratch. A privileged RL teacher~\citep{viral2025} and a procedural task plan are thus alternative ways to encode a task; a head-to-head comparison is left to future work.
\vspace{-1em}
\paragraph{Real-to-sim-to-real and photorealistic simulation.}
Building simulators from real-world captures is an established route to robot learning. RialTo~\citep{torne2024rialto} robustifies imitation policies learned from a few real demonstrations with RL in scanned digital twins, PolaRiS~\citep{jain2025polaris} turns short video scans into simulated environments for scalable policy evaluation, DRAWER~\citep{xia2025drawer} reconstructs articulated interactive scenes, and PhysTwin~\citep{jiang2025phystwin} recovers deformable-object physics from video. Earlier sim-to-real work instead randomizes non-photorealistic renderers~\citep{dr2017,sun2024arch}, and mesh reconstruction via SAM3D~\citep{sam3d2024} scales synthetic data but inherits the mesh-rendering gap; we use it as our non-3DGS baseline.
3D Gaussian Splatting~\citep{kerbl3Dgaussians} has been used as a simulator backbone for fixed-base manipulation~\citep{gsworld2025,shorinwa2024splat}, high-throughput RL on locomotion and tabletop tasks~\citep{gaussgym2025,gsplayground2026}, visuomotor drone navigation~\citep{sousvide2025, adang2025singeronboardgeneralistvisionlanguage, gradnav2025, gradnavpp2025}, and aerial manipulation~\citep{tucker2026pimakeflyphysicsguided}. Whereas these works target manipulation, locomotion, navigation, or evaluation, \ours applies real-to-sim to teleop-free demonstration generation for fine-tuning pretrained VLAs on humanoid loco-manipulation, where decoupling recorded motion from appearance lets one motion dataset be re-rendered for new target appearances.

\begin{figure*}[t]
  \centering
  \includegraphics[width=\linewidth]{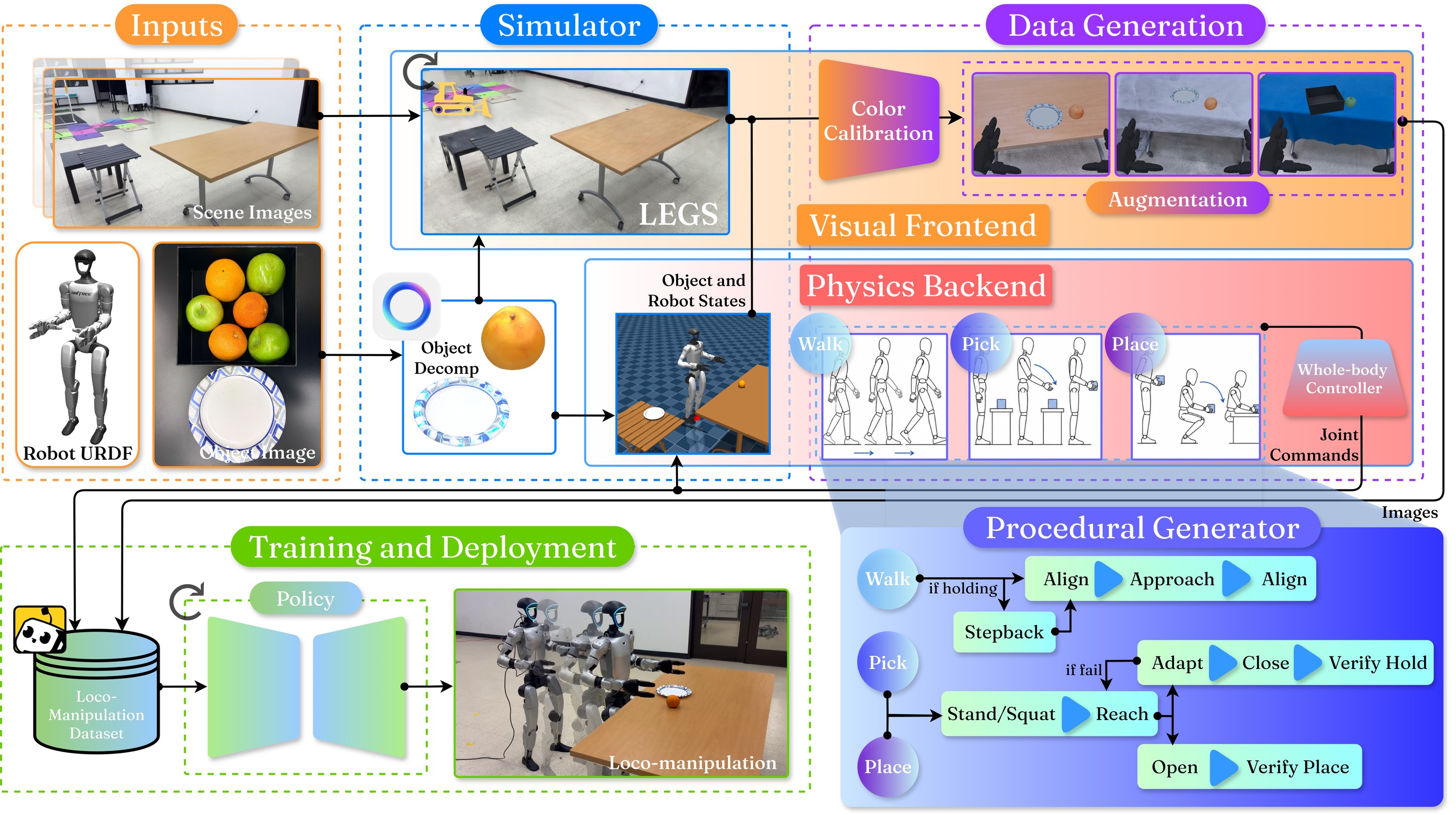}
\caption{
  \textbf{The \ours pipeline.}
  A scene video and object photo are reconstructed into a 3DGS background and SAM3D meshes, which feed the \ours simulator.
  The simulator decouples a \textbf{visual frontend} (3DGS and mesh compositor with color calibration) from a \textbf{physics backend} (MuJoCo with the low-level whole-body controller).
  A procedural generator produces labeled demonstrations, re-rendered under scene and object augmentations, and used to fine-tune a VLA backbone (\psizero, \pihalfpoint, \groot) for real-robot deployment.
  }
  \label{fig:overview}
\end{figure*}

\section{The \ours Framework}
\label{sec:legs}

The framework is organized around two decouplings. \emph{Physics from rendering}: a MuJoCo~\citep{mujoco2012} backend running a low-level whole-body controller resolves dynamics on mesh geometry, while a separate rendering frontend produces image observations. \emph{Static background from dynamic foreground}: the background is a fixed \gsplat field rendered for appearance only, the foreground is rasterized meshes that move under physics, and layers are composited per frame by a depth test. Motion recorded against the foreground meshes is therefore independent of scene appearance, so a single motion dataset can be re-rendered under swapped backgrounds and meshes (Figure~\ref{fig:overview}).

\subsection{Visual Frontend}
\label{sec:visual_frontend}

\paragraph{Background and foreground.}
Each scene is captured with a $\sim$1--2~minute handheld video and reconstructed into a \gsplat field via COLMAP~\citep{schoenberger2016sfm} poses followed by \gsplat optimization (Appendix~\ref{app:sim}). Because \gsplat minimizes rendering error against real photographs, novel views are photorealistic, reducing the visual sim-to-real gap relative to mesh rendering.
The foreground meshes come from the robot URDF and SAM3D~\citep{sam3d2024} reconstructions of objects and props from a single handheld image. At each frame the foreground is rasterized and depth-composited with the background.

\begin{figure}[t]
  \centering
  \includegraphics[width=0.87\linewidth]{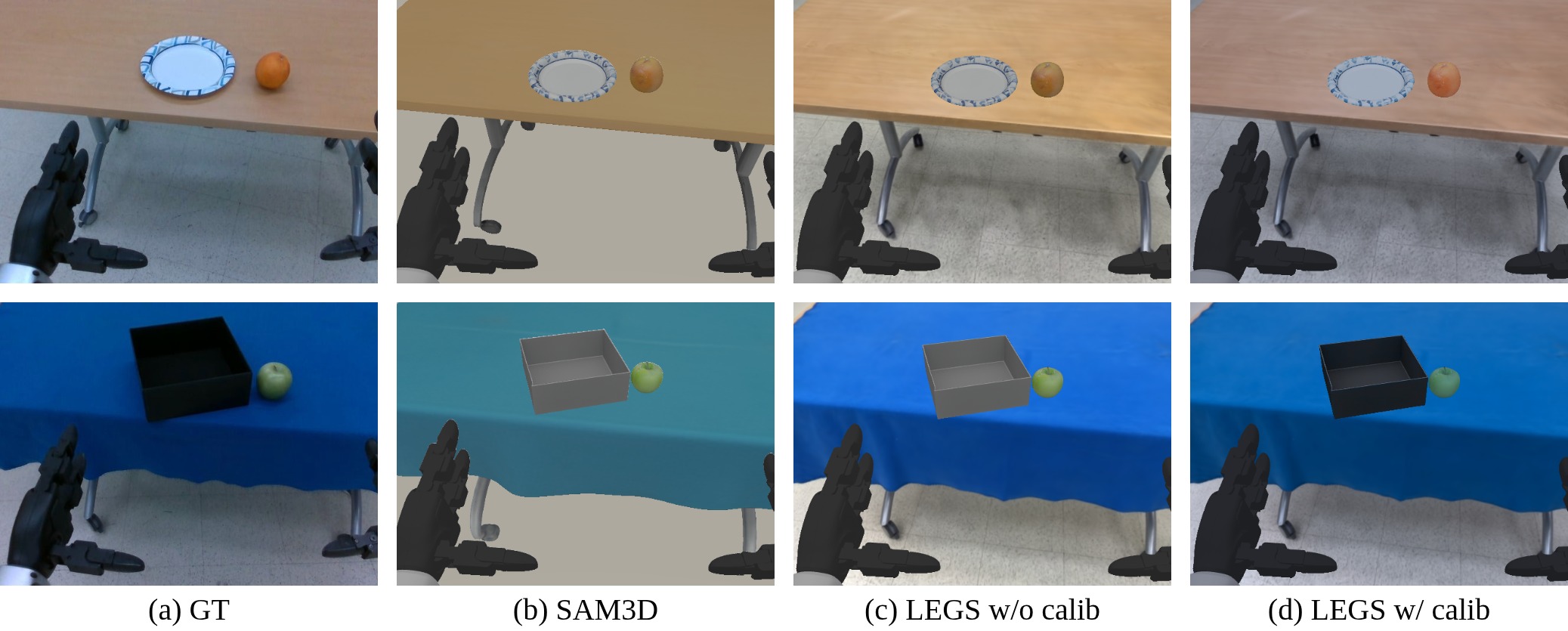}
\caption{
    \textbf{Egocentric views of the same task moment.}
    Columns: (a) real on-robot image, (b) SAM3D mesh-only baseline, (c, d) raw and color calibrated \ours renders.
    The second row shows the \emph{same recorded motion} re-rendered under swapped object and scene meshes.
}
  \label{fig:rendering}
  \vspace{-1em}
\end{figure}

\paragraph{Color calibration.}
The composite mixes two color spaces, the scanning device's ISP (\gsplat) and SAM3D's regressed albedo (meshes), which must be mapped to a third, the on-robot deployment camera. We disable auto-exposure and auto-white-balance on both cameras so the mapping is stationary. We close the gap in two stages:
\begin{equation*}
c_d \approx M\,[R(c_m) \sqcup c_b],
\end{equation*}
where $c_m$ is a mesh vertex color, $c_b$ a background \gsplat pixel, $c_d$ the deployment-camera pixel, $R(\cdot)$ a per-mesh diagonal scale in linear RGB matching each mesh's rendered mean to the handheld scan inside its SAM2~\citep{ravi2025sam} mask, $\sqcup$ depth-test compositing, and $M \in \mathbb{R}^{3\times 3}$ a global color-correction matrix fit by least squares from a 24-patch ColorChecker photographed by both cameras. As Figure~\ref{fig:rendering} shows, the per-mesh scale $R$ corrects SAM3D's mis-regressed albedo---the box that SAM3D renders as washed-out grey is restored to black---while the global matrix $M$ shifts the \gsplat background (table and floor) toward the deployment camera, together bringing the composite close to the real on-robot image. Derivations are in Appendix~\ref{app:calibration}.

\subsection{Physics Backend and Controller Interface}
\label{sec:physics_backend}
 
The physics backend resolves dynamics on mesh geometry in MuJoCo at 500~Hz, separately from the rendering pipeline. The dynamic foreground reuses the SAM3D object meshes and URDF robot model from Section~\ref{sec:visual_frontend}, with object collision geometry decomposed via CoACD~\citep{coacd2022}. The static background has its own collision mesh, derived from a SAM3D reconstruction aligned to the \gsplat field.

We use SONIC~\citep{sonic2025} as the low-level whole-body controller, an RL-trained policy that handles whole-body coordination from a high-level command interface. SONIC accepts an 18-D command shared between simulation and the real robot: a 6-D wrist SE(3) pose plus a continuous grip scalar per arm (14-D upper-body), concatenated with a 4-D base command ($v_x, v_y, \omega_z, h$). The same SONIC binary runs in simulation and on the robot, so deployment requires no changes to the deployed policy.

\subsection{Procedural Episode Generation}
\label{sec:primitives}

\ours leverages the privileged information (robot, object poses) inherent to a simulator to procedurally generate episodes, further reducing human labor. \ours composes parametrized motion primitives whose arguments are scene-level (target pose, object identity, arm choice) rather than joint-level, which lets a single task plan produce a different valid trajectory under each randomized initial condition. Each pick-and-place task decomposes into three high-level motions (\texttt{Walk}, \texttt{Pick}, \texttt{Place}), each composed of lower-level primitives (Figure~\ref{fig:overview}). Each motion ends with a verification check, and only verified successful episodes are saved (85--94\% of generated episodes across tasks). Randomization ranges, the full primitive protocol, and the split between shared and per-task specification are in Appendices~\ref{app:randomization} and~\ref{app:data}.

The recorded 18-D command stream is independent of the rendering inputs, so the resulting dataset can be re-rendered under swapped backgrounds and meshes at GPU cost only, with no new motion collection (Section~\ref{sec:gen_results}).

\section{Experiments}
\label{sec:experiments}

We evaluate \ours on three pick-and-place tasks of increasing whole-body difficulty, fine-tuning three pretrained VLA backbones (\psizero, \pihalfpoint, \groot) on each of three data conditions, including human teleoperation and a mesh-only synthetic baseline.
 
\subsection{Experimental Setup}
\label{sec:setup}

\begin{figure}[h]
    \centering
    \includegraphics[trim={15cm 0cm 0cm 7cm}, clip, width=\linewidth]{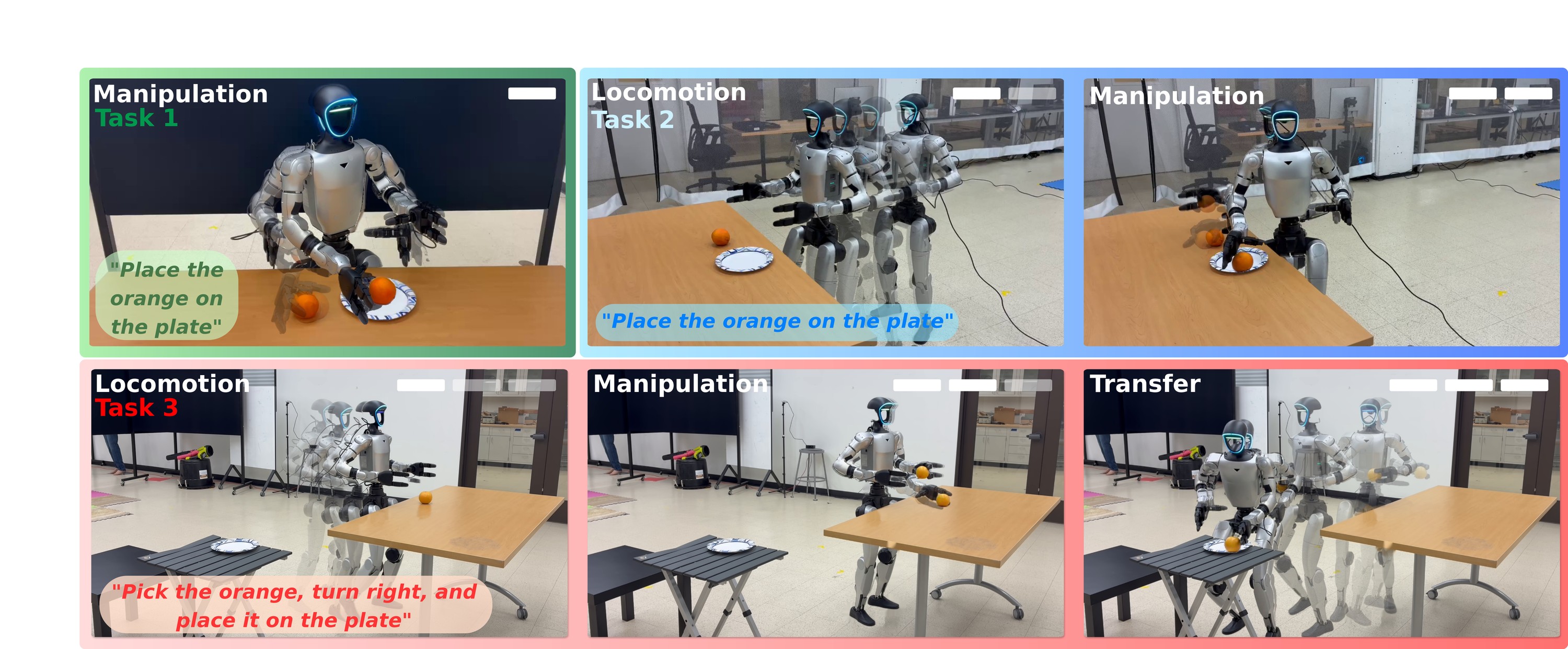}
    \caption{
      \textbf{Three pick-and-place tasks of increasing whole-body difficulty.}
      \textcolor{task1}{\textbf{Task~1 (manipulation-only)}}: pick and place the orange on the plate without locomotion.
      \textcolor{task2}{\textbf{Task~2 (loco-manipulation, easy)}}: walk to the table, then pick and place.
      \textcolor{task3}{\textbf{Task~3 (loco-manipulation, hard)}}: walk, pick, turn, walk to a low table, squat, and place.
    }
    \label{fig:tasks}
\end{figure}

\paragraph{Tasks and protocol.}
The three tasks (Figure~\ref{fig:tasks}) are all pick-and-place on the Unitree~G1. Across all tasks the robot picks an orange and places it on a white plate sitting on a wood-grain table. Task~1 and Task~2 share the language prompt \emph{``place the orange on the plate''}, and Task~3 uses \emph{``pick the orange, turn right, and place it on the plate''}. Figure~\ref{fig:task3_frames} (Appendix~\ref{app:task3}) shows the full Task~3 trajectory annotated with its subtask stages and motion primitives, alongside the egocentric views for each data condition. A trial can fail at any subtask, so we report end task success rate (TSR) in the main text and stage-wise cumulative success in Appendix~\ref{app:stagewise}.
The robot has a 29-DoF body and 7-DoF Dex3 hand per side, with a single head-mounted Intel RealSense~D435 streaming RGB at 30~Hz.
We run $R=10$ trials per data condition, backbone, and task, with initial conditions drawn independently per trial: the orange and plate positions are perturbed by $\pm5$~cm and the robot heading by $\pm10^\circ$, plus the robot base position by $\pm10$~cm on the loco-manipulation tasks. On the long-horizon Task~3 we extend this to 20 trials per cell in Appendix~\ref{app:matched}.

\paragraph{Data conditions.}
We compare three data conditions.
\textbf{Teleop~(50)} collects 50 real demonstrations per task, recorded by a non-expert operator via SONIC's VR 3-point teleoperation interface~\citep{sonic2025}.
\textbf{SAM3D~(200)} is a mesh-only synthetic baseline identical to \ours~(200) except that the \gsplat background is replaced by SAM3D meshes and color calibration is disabled (Appendix~\ref{app:sam3d}).
\textbf{\ours~(200)} is our main dataset, produced by the procedural generator of Section~\ref{sec:primitives} and rendered as mesh foreground composited over the \gsplat background with color calibration enabled.
Because color calibration assumes a fixed camera ISP, \ours policies are deployed with fixed exposure and white balance, while Teleop and SAM3D use the camera's default automatic mode; Appendix~\ref{app:matched} removes this confound by re-evaluating \ours under the automatic mode.
We also report \textbf{\ours~(50)}, the first 50 episodes of \ours~(200), for a budget-matched comparison against teleop.
The appearance-adaptation experiment of Section~\ref{sec:gen_results} runs on the strongest in-distribution backbone per task from Table~\ref{tab:main_tsr}, \psizero for Tasks~1--2 and \groot for Task~3.
\paragraph{Foundation-model fine-tuning.}
We fine-tune three VLAs from their released checkpoints (\psizero~\citep{psi0}, \pihalfpoint~\citep{pi05}, \groot~\citep{grootn12024}), all sharing SONIC as the controller and the 18-D command of Section~\ref{sec:physics_backend}. Within each backbone the fine-tuning recipe is fixed across data conditions, so within-backbone comparisons isolate the data source. Recipes differ across backbones (full fine-tuning for \pihalfpoint, action-head-only for the others). Full per-backbone details are in Appendix~\ref{app:training}.
 
\subsection{Performance Across Three VLA Backbones}
\label{sec:main_results}

\paragraph{\ours matches or exceeds teleop.}
\ours~(200) is best or tied on all nine (backbone, task) experiments of Table~\ref{tab:main_tsr}, and the gap over Teleop~(50) widens with task difficulty: modest on Task~1, larger on Task~2, and decisive on Task~3 where Teleop~(50) collapses to 0/10 across all backbones while \ours~(200) achieves 2--6/10. The ranking holds across all three backbones despite different fine-tuning recipes, suggesting that the gain is a property of the data pipeline, not any one backbone. Individual 10-trial cells are underpowered, so we draw conclusions from aggregates: across the nine experiments, \ours~(200) reaches 60/90 (66.7\%, Wilson 95\% CI 56.4--75.5) versus 22/90 for Teleop~(50) (24.4\%, 16.7--34.2) and 30/90 for SAM3D~(200) (33.3\%, 24.5--43.6).

\paragraph{The gain is not a budget artifact.}
The budget-matched \ours~(50) condition matches or beats Teleop~(50) on every experiment, including Task~3 where teleop produces zero successes, indicating that the improvement is not explained by \ours's larger 200-demonstration budget alone. The further gap from \ours~(50) to \ours~(200) is consistent with standard imitation-learning data scaling. Appendix~\ref{app:stagewise} shows that this gap appears at the \texttt{pick} and \texttt{place} stages, the precise close-range manipulation stages where extra demonstrations help most, while the methods stay comparable at \texttt{walk}.

 \begin{table}[t]
\centering
\small
\setlength{\tabcolsep}{4pt}
\caption{
  \textbf{End task success out of 10 trials, per data condition, backbone, and task.}
  \ours~(200) is our main dataset. \ours~(50) is the first 50 episodes of \ours~(200), for a budget-matched comparison against the 50-demo teleop baseline.
  Best per (backbone, task) in \textbf{bold}. Stage-wise cumulative successes for the same experiments are shown in Appendix~\ref{app:stagewise}.
}
\label{tab:main_tsr}
\begin{tabular}{l ccc c ccc c ccc}
\toprule
& \multicolumn{3}{c}{\textbf{Task 1}} & & \multicolumn{3}{c}{\textbf{Task 2}} & & \multicolumn{3}{c}{\textbf{Task 3}} \\
\cmidrule(lr){2-4} \cmidrule(lr){6-8} \cmidrule(lr){10-12}
Data condition & $\psi_0$ & $\pi_{0.5}$ & \groot & & $\psi_0$ & $\pi_{0.5}$ & \groot & & $\psi_0$ & $\pi_{0.5}$ & \groot \\
\midrule
Teleop (50)         & 8/10 & 2/10 & 4/10 & & 6/10 & 0/10 & 2/10 & & 0/10 & 0/10 & 0/10 \\
SAM3D (200)         & 7/10 & 1/10 & 6/10 & & 4/10 & 3/10 & 5/10 & & 1/10 & 0/10 & 3/10 \\
\ours (50)          & 8/10 & 5/10 & \textbf{9/10} & & 8/10 & 4/10 & 7/10 & & 3/10 & 1/10 & 5/10 \\
\ours (200)  & \textbf{10/10} & \textbf{6/10} & \textbf{9/10} & & \textbf{9/10} & \textbf{5/10} & \textbf{8/10} & & \textbf{5/10} & \textbf{2/10} & \textbf{6/10} \\
\bottomrule
\end{tabular}
\end{table}

\subsection{Effect of the Rendering Frontend}
\label{sec:photorealism}

The SAM3D~(200) condition shares the \ours motion data and pipeline but replaces the \gsplat background with SAM3D meshes and disables color calibration. SAM3D~(200) lies strictly below \ours~(200) on every experiment of Table~\ref{tab:main_tsr}: it averages 33\% end task success against 67\% for \ours~(200), a 33-point mean drop that roughly halves the success rate. SAM3D retains success comparable to \ours at \texttt{walk}, but diverges from \ours during \texttt{pick} and \texttt{place} on every task and backbone. Because the two conditions differ in background representation, color calibration, and deployment camera mode, this comparison establishes a frontend-level effect of calibrated \gsplat rendering over an untuned mesh renderer, not the contribution of \gsplat alone.
To remove the camera confound, Appendix~\ref{app:matched} re-evaluates the Task~3 \ours checkpoints with the camera in the same automatic mode as the baselines ($n{=}20$ per cell): \ours~(200) changes from 23/60 to 18/60 ($p{=}0.41$), still ahead of SAM3D~(200) at 6/60 ($p<0.01$, stratified by backbone). Separating the \gsplat background from color calibration requires retraining and is left to future work.
\subsection{Low-Labor Adaptation to Target Scene Appearance}
\label{sec:gen_results}

\begin{figure}[t]
\centering
\includegraphics[width=\linewidth]{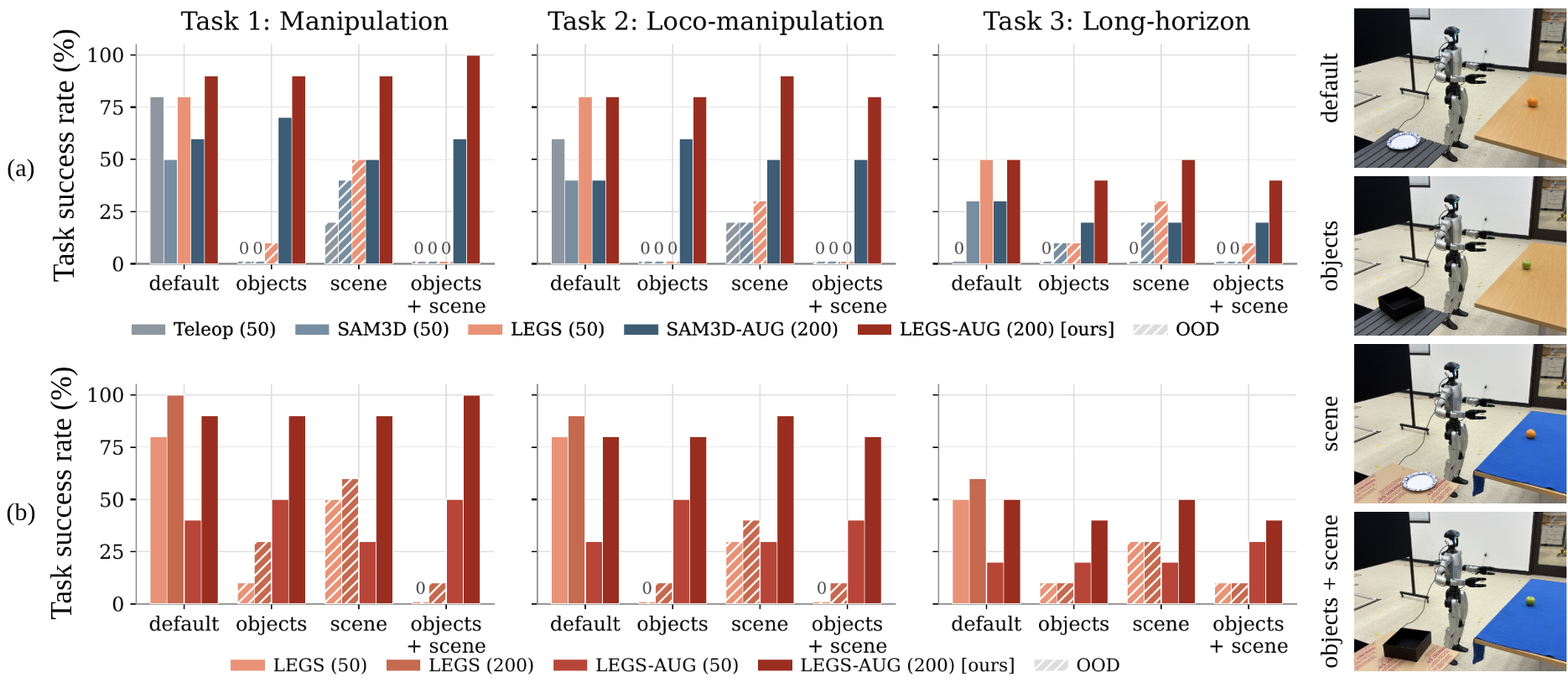}
\caption{
  \textbf{Real-robot TSR (\%) under appearance shift.}
  Variations cover \emph{default}, \emph{objects} (apple and box), \emph{scene} (blue table), and \emph{objects + scene}, shown as third-person views on the right.
  Hatched bars are out-of-distribution for the training set, solid bars are in-distribution.
  Tasks~1--2 are evaluated on \psizero, Task~3 on \groot.
  \textbf{(a)} Comparison across data sources at the 50-episode budget against re-rendered \textsc{SAM3D-aug~(200)} and \textsc{LEGS-aug~(200)}.
  \textbf{(b)} Within-\ours ablation isolating dataset size from re-rendering.
}
\label{fig:variations_results}
\end{figure}

\ours's appearance-from-rendering decoupling (Section~\ref{sec:primitives}) predicts that adapting to a new scene's appearance should require only re-rendering recorded motion, not collecting new motion. We test this on four appearance conditions, varying objects and scene independently (Figure~\ref{fig:variations_results}), measuring both per-condition collection cost and real-robot TSR.

\paragraph{Appearance and prompt swap.}
The \emph{default} condition uses the orange-and-plate setup and prompts of Section~\ref{sec:setup}. The \emph{scene} condition swaps only the table for a blue table and keeps the default prompt. The \emph{objects} and \emph{objects + scene} conditions further swap the orange for an apple and the plate for a box, with the prompt updated to \emph{``put the apple in the box''} (Tasks~1--2) or \emph{``pick the apple, turn right, and put it in the box''} (Task~3). For \textsc{LEGS-aug} and \textsc{SAM3D-aug}, training episodes are re-rendered with the new meshes and re-labeled with the new prompt (composition in Appendix~\ref{app:randomization}). Because the augmented training set contains the exact evaluation scene, object meshes, and prompts, this experiment measures target-specific adaptation, not generalization to unseen assets. Teleop, SAM3D~(50), and \ours~(50) are trained on the default data only and see the shifted visuals and prompt out of distribution at test time. Teleop cannot be re-rendered, and the un-augmented synthetic baselines are matched to teleop's training budget for fairness.

\paragraph{Cost.}
Producing 50 Task~3 demonstrations takes 1.5 operator-hours for teleop and roughly 0.5 GPU-hours for \ours on a single RTX~4090, with no operator involvement. Each additional appearance condition then costs more than 1.5 operator-hours for teleop, which must re-collect motion in the new scene, versus roughly 0.1 GPU-hours for \ours (and $\sim$0.2 for the SAM3D baseline), which only re-render the existing motion (Appendix~\ref{app:dataset_cost}); a new scene additionally needs a 1--2~minute handheld video and unattended \gsplat reconstruction. Adapting to a target scene therefore shifts from an operator-bound to a compute-bound problem, and the compute is parallelizable across GPUs.

\paragraph{Re-rendering enables adaptation; the calibrated \gsplat frontend improves it.}
Without re-rendering, all three 50-episode baselines (Teleop~(50), SAM3D~(50), and \ours~(50)) collapse to 0--1/10 on \emph{objects} and \emph{objects + scene} across all three tasks, confirming that the joint appearance-and-prompt shift is genuine and that no method generalizes to it from default data alone. With re-rendering, both \textsc{SAM3D-aug~(200)} and \textsc{LEGS-aug~(200)} retain success on every shifted appearance condition (Figure~\ref{fig:variations_results}a), with \textsc{LEGS-aug} consistently ahead: on the hardest \emph{objects + scene} shift it reaches 100\%, 80\%, and 40\% on Tasks~1--3, versus 60\%, 50\%, and 20\% for \textsc{SAM3D-aug}. Teleop cannot participate in this gain because a physically captured dataset cannot be re-rendered.

\paragraph{Augmentation beats scale under object-and-prompt shift.}
Figure~\ref{fig:variations_results}(b) isolates dataset size from augmentation within \ours: on the \emph{objects} and \emph{objects + scene} shifts, re-rendering into the new object and prompt distribution (\textsc{LEGS-aug~(50)}) beats not only \textsc{LEGS~(50)} but also the fourfold-larger \ours~(200) (on \emph{objects + scene} it reaches 50\%, 40\%, and 30\% on Tasks~1--3 while \ours~(200) stays at 10\% across all tasks) because the default-only policy never sees the new object or prompt. On the milder \emph{scene}-only shift, the larger default dataset is comparable or better, so re-rendering pays off specifically when the shift changes what the policy must recognize, not merely the background.

\section{Limitations}
\label{sec:limitations}

\ours removes teleoperation but not task engineering: each task needs a plan over shared primitives, and new task families such as articulated-object interaction or tool use would need new primitives (insertion would also need force control). The one-time cost of building the framework is not included in our cost comparisons. All three evaluated tasks are structured pick-and-place, and our attribution of the gain over the mesh baseline is to the calibrated \gsplat frontend as a whole; a factorial ablation of background and calibration, and comparisons against a tuned, photometrically randomized mesh renderer, target-condition teleoperation, or a privileged RL teacher, are left to future work.
Each cell uses 10 trials (20 on Task~3 in Appendix~\ref{app:matched}) and a single training seed, so per-cell differences are not individually conclusive, and we do not report a taxonomy of falls, collisions, or safety interventions during evaluation. \textsc{LEGS-aug} adapts to specified target appearances rather than generalizing to unseen ones.
Our color calibration disables auto-exposure and auto-white-balance to keep the mapping stationary, so the trained policy is sensitive to lighting changes outside the distribution captured during the original \gsplat scan. Adapting to a new scene still requires a 1--2~minute handheld video, which is cheaper than teleop re-collection but not free. The \gsplat reconstruction degrades on highly reflective or transparent surfaces and assumes a static background, so scenes with moving humans or other dynamic content would require dynamic-\gsplat methods that we do not implement. Our procedural generator produces feasible motion but does not optimize for dynamic efficiency, unlike approaches based on trajectory optimization or MPC. The appearance-adaptation suite varies only small graspable objects of similar size and geometry, so transfer to substantially different object categories is not tested. We evaluate on a single robot platform (Unitree~G1), camera (Intel RealSense~D435), and whole-body controller (SONIC); generalization across humanoid hardware is not tested. Finally, scanning real deployment spaces may capture private information and should be done with appropriate consent.

\section{Conclusion}
\label{sec:conclusion}

We presented \ours, a hybrid simulator that composites a mesh foreground over a photorealistic \gsplat background, together with a procedural generator, producing labeled humanoid loco-manipulation demonstrations without any teleoperation, seed demonstration, or human video. Across three pick-and-place tasks and three pretrained VLA backbones on a Unitree~G1, \ours data matches or exceeds human teleoperation on every experiment, outperforms a mesh-only synthetic baseline without the calibrated \gsplat frontend, and adapts to specified object-and-scene appearance targets at a fraction of the operator cost. For this task family, calibrated \gsplat rendering and procedural motion generation can replace humanoid teleoperation at the data-collection step.

\clearpage
\acknowledgments{This work was supported in part by ONR grant N00014-23-1-2354 and NSF grant 2342246. Timothy Chen is supported by a NASA NSTGRO fellowship. We are grateful for this support.}

\bibliography{reference}

\newpage
 
\appendix

\section{Simulator Implementation Details}
\label{app:sim}

\subsection{Reconstruction Pipeline}
Background reconstruction: (1) iPhone video capture (handheld, $\sim$1--2~minute sweep of the room), (2) COLMAP sparse reconstruction for camera pose estimation, (3) 3DGS training through nerfstudio (30K iterations, single workstation GPU, $\sim$10--15~min). Alternatively, one can use the many image-to-\gsplat apps (e.g. Polycam).

For the background collision mesh used by physics, we reconstruct a mesh with SAM3D~\citep{sam3d2024} and align it to the \gsplat field via the shared COLMAP poses. Some image-to-\gsplat apps (e.g. Kiri Engine) instead export a mesh in the same frame as the \gsplat reconstruction, skipping this alignment.

\paragraph{Reconstruction robustness.} Because object stability depends on the reconstructed support surfaces, we checked their geometry and the resulting contact behavior. Across ten independent SAM3D reconstructions of each Task~3 table, the solid tabletops are planar to within $1^\circ$. In 400 simulated placements of the task objects at task poses, no object slid or fell.

\subsection{Front-End / Back-End Interface}

As noted in Section~\ref{sec:physics_backend}, the simulator front-end and the SONIC controller back-end use identical interfaces in simulation and on the real robot. Concretely, they communicate over ROS/DDS pub-sub: the front-end publishes the 18-dim command stream and receives proprioceptive state, and the same DDS topic names are reused across deploy targets. For interactive visualization and scene inspection during episode generation, the front-end uses viser~\citep{yi2025viser}.

\section{Stage-Wise Cumulative Success}
\label{app:stagewise}

\begin{figure}[h]
\centering
\includegraphics[width=\linewidth]{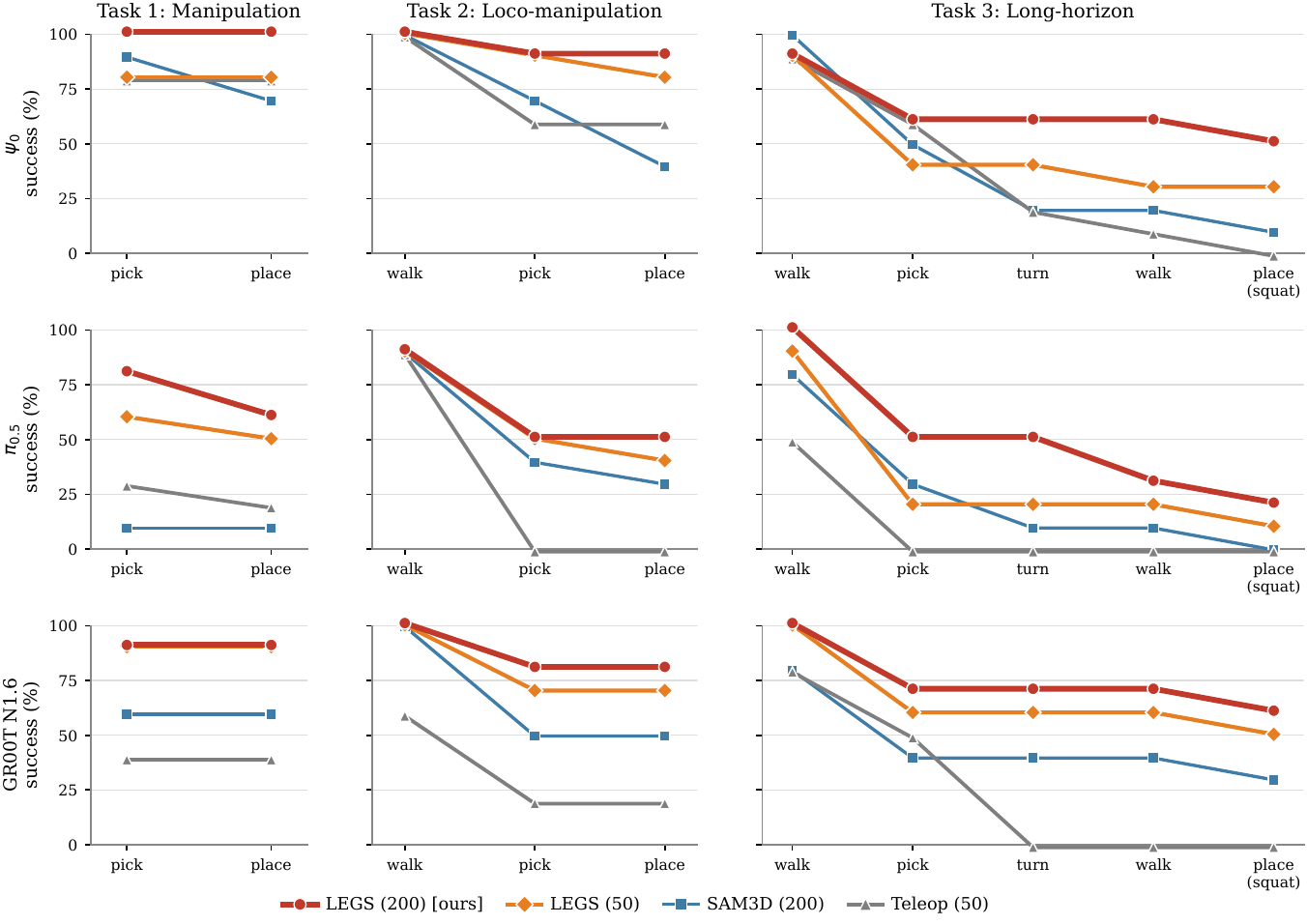}
\caption{
  \textbf{Stage-wise cumulative success, faceted by task (columns) and backbone (rows).}
  Each panel shows four data conditions (\ours~(200) [ours], \ours~(50), SAM3D~(200), Teleop~(50)) tracing the percentage of 10 trials still successful through each subtask stage; the final stage equals overall TSR.
  Small per-condition y-offsets keep overlapping lines visible.
}
\label{fig:main_results}
\end{figure}

Each task in Section~\ref{sec:setup} is a sequence of subtasks, so a trial can fail at any stage, and the stage at which a method fails is itself diagnostic.
Figure~\ref{fig:main_results} reports the per-stage cumulative success underlying the end-TSR numbers of Table~\ref{tab:main_tsr}.
Across most cells the methods are comparable at the \texttt{walk} stage but separate during \texttt{pick} and \texttt{place}, the camera regime in which the egocentric view moves close to the object and rendering fidelity dominates. Teleop typically collapses toward zero at these stages, while SAM3D plateaus at a lower success level than \ours; \ours retains the highest success through the manipulation stages on every backbone.

\section{Color Calibration Pipeline}
\label{app:calibration}
The composited observation mixes three color spaces: the iPhone ISP output (\gsplat), SAM3D's regressed albedo (meshes), and the RealSense~D435 ISP output (the deployment camera). To keep the mapping stationary, we disable auto-exposure and auto-white-balance on both cameras. This only applies to \ours; SAM3D and teleop apply no color calibration and run the RealSense in its default auto-exposure/auto-white-balance mode. We report the deployment RealSense~D435 settings (exposure 10~ms, white balance 3000~K), and the iPhone scan settings are locked but absorbed into the Stage-2 calibration. We then align the three spaces in two stages.

\paragraph{Stage 1: mesh-to-deployment albedo match.}
The SAM3D vertex colors approximate the input texture but are systematically muted, as the network regresses surface albedo rather than the ISP tone curve of the scanning device, and the mesh is rendered under physically based (PBR) lighting that further modulates the baked colors. A chart-based calibration does not apply, since no chart is visible in an arbitrary object photograph. We instead match the rendered object statistics to the source photograph directly. For each mesh we render an interior view under the deployment renderer's PBR lighting (directional headlight intensity~1.0, ambient~0.3) and take $\mu_S^{(c)}$, the per-channel mean of linear RGB over pixels with $\alpha > 200$. We pair it with $\mu_T^{(c)}$, the per-channel mean of linear RGB over iPhone pixels inside the SAM2 mask. The diagonal scale
\begin{equation*}
s^{(c)} \;=\; \mathrm{clip}\!\left(\frac{\mu_T^{(c)}}{\mu_S^{(c)}},\; 0.5,\; 3.0\right)
\end{equation*}
is multiplied into the mesh's linear vertex colors, which are then clipped to $[0,1]$ and re-exported; geometry is unchanged. Operating diagonally in linear RGB rather than as a full $3{\times}3$ keeps the multiplier well-conditioned: a full-covariance fit needed entries above~6 to match the iPhone's wider intra-region variance, which clipped vertex colors and shifted hue. Restricting the target statistics to the masked region is essential, since background pixels would bias the match toward scene illumination rather than the object. Two conventions matter: the scale is fit against the PBR-lit render used at data-generation time, not an idealized unlit appearance; and since glTF~2.0 stores vertex colors as linear RGB, the calibrated colors are written in linear RGB so the renderer's gamma-encode recovers the intended sRGB appearance without double-encoding.

\paragraph{Stage 2: iPhone-to-RealSense color-correction matrix.}
With all assets in iPhone space, a single $3{\times}3$ matrix maps composited frames to RealSense space. We photograph a 24-patch color chart with both cameras under controlled illumination, extract the median sRGB value of each patch, and linearize via the sRGB EOTF. Stacking the linear RealSense and iPhone patch colors as $P,Q\in\mathbb{R}^{24\times3}$, the matrix is the least-squares solution
\begin{equation*}
M^\star \;=\; \arg\min_M \|P - QM\|_F^2 \;=\; (Q^\top Q)^{-1}Q^\top P .
\end{equation*}
At render time, each composited frame is linearized, multiplied by $M^\star$, clipped to $[0,1]$, and re-encoded to sRGB. Applying $M^\star$ at the composite level rather than per object keeps meshes and the \gsplat background spatially consistent with no boundary artifacts.

\section{Kinematic Randomization Parameters}
\label{app:randomization}
Per-reset randomization ranges, sampled uniformly. ``Object'' ranges apply to every randomized object in the scene (the picked object and the placement target).

\begin{table}[h]
\centering
\small
\begin{tabular}{lll}
\toprule
Quantity & Task 1 (manipulation) & Tasks 2--3 (loco-manip) \\
\midrule
Robot $x, y$ offset (cm)   & 0         & $\pm 10$ \\
Robot heading (deg)        & $\pm 10$   & $\pm 10$ \\
Object $x, y$ offset (cm)  & $\pm 5$   & $\pm 5$ \\
Object yaw (deg)           & $\pm 180$ & $\pm 180$ \\
\bottomrule
\end{tabular}
\end{table}

For the appearance-condition evaluation (Section~\ref{sec:gen_results}), the objects and tables are swapped for the target-appearance assets (orange and plate, apple and box, wood or blue table) and the episodes are re-rendered; no photometric augmentation is applied. \textsc{LEGS-aug~(50)} allocates 12/13/12/13 episodes to the \emph{default}, \emph{objects}, \emph{scene}, and \emph{objects + scene} conditions, and \textsc{LEGS-aug~(200)} uses 50 episodes per condition.

\section{SAM3D Synthetic Baseline Details}
\label{app:sam3d}
\paragraph{Reconstruction.}
We reconstruct each object (orange, plate, tables, and the swap assets used in Section~\ref{sec:gen_results}) from a single handheld image using SAM3D~\citep{sam3d2024} and extract textured meshes; the robot is loaded from its manufacturer URDF, as in the \ours condition.

\paragraph{Renderer.}
A standard mesh rasterizer renders the entire scene (foreground objects plus floor, walls, and props), with no \gsplat background and color calibration disabled. We did \emph{not} tune the renderer to defeat the visual gap; this matches a typical mesh-synthetic pipeline a practitioner would build.

\paragraph{Otherwise identical.}
The procedural episode generator, MuJoCo physics, CoACD collision decomposition, SONIC WBC backend, action and state spaces, per-reset kinematic randomization, and the 200-demo budget are all held identical to the \ours condition, so differences are attributable to the rendering frontend (\gsplat background and color calibration) and the deployment camera mode rather than data scale; Appendix~\ref{app:matched} removes the camera-mode difference. No photometric augmentation is applied.

\section{Dataset-Collection Cost}
\label{app:dataset_cost}
\begin{table}[h]
\centering
\small
\setlength{\tabcolsep}{5pt}
\caption{
  \textbf{Dataset-collection cost} (wall-clock hours, Task~3, 50 episodes each).
  \emph{Initial} is the cost of the first 50-episode dataset; the right column is the marginal cost of one additional appearance condition.
}
\label{tab:dataset_cost}
\begin{tabular}{l c c}
\toprule
Dataset & Initial cost (hr) & + 1 appearance condition (hr) \\
\midrule
Teleop (50)  & 1.5 (operator) & $>$1.5 (operator + hardware fatigue) \\
SAM3D (50)   & 0.5 (compute)  & 0.2 (re-render only) \\
\ours (50)   & 0.5 (compute)  & 0.1 (re-render only) \\
\bottomrule
\end{tabular}
\end{table}

Teleop~(50) is one operator collecting 50 demonstrations of Task~3, the most expensive task at $\sim$30~s per episode. The synthetic conditions instead run the procedural generator at real-time physics for initial motion capture, then re-render the recorded motion on a single RTX~4090 with +20 CPU cores for each additional appearance condition. \ours~renders mesh foreground over a rasterized \gsplat background, while SAM3D~re-shades the full mesh scene (foreground plus background props), making its re-render slightly more expensive.

The table makes two asymmetries visible.
\emph{(i)~Initial dataset}: both synthetic conditions produce 50 demonstrations in $3\times$ less wall-clock time than teleop and with zero operator hours, since the procedural generator runs headlessly and shares one motion dataset.
\emph{(ii)~Per-condition scaling}: each additional appearance condition costs only GPU time for the synthetic conditions ($\sim$0.1~hr for \ours, $\sim$0.2~hr for SAM3D) but more than 1.5~hr of operator work for teleop, whose motion is bound to the capture scene and must be re-collected.
The second asymmetry is what makes covering many appearance conditions practical in simulation and impractical for teleop; the \ours-vs-SAM3D gap further shows the \gsplat background is cheaper to re-render than a full mesh scene.

\section{Data Collection Protocol}
\label{app:data}

\subsection{Teleoperation Setup}

\begin{wrapfigure}{r}{0.3\linewidth}
  \centering
  \vspace{-\baselineskip}
  \includegraphics[trim={0 3cm 0 1cm}, clip, width=\linewidth]{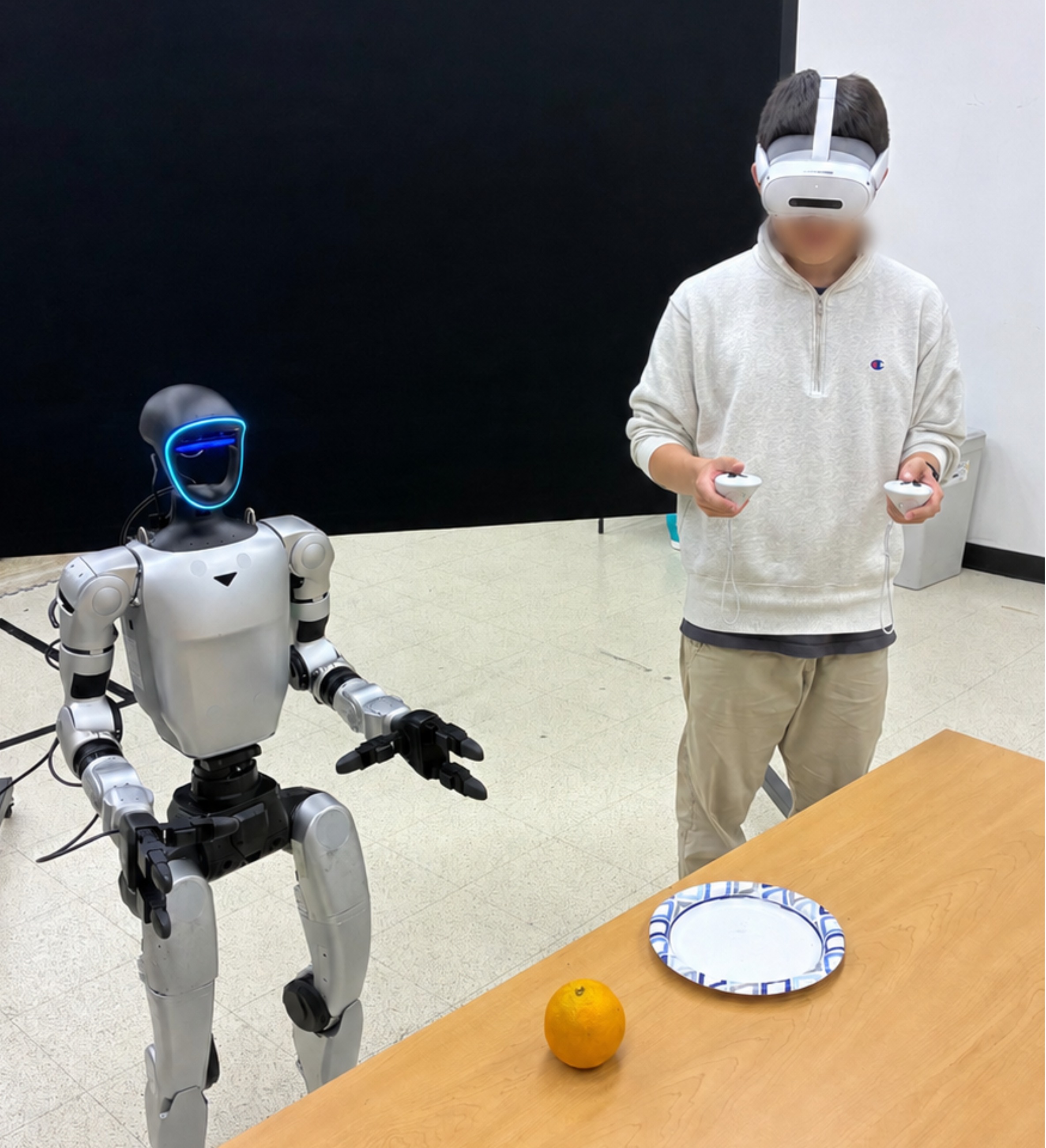}
  \caption{VR three-point teleoperation setup.}
  \label{fig:teleop}
\end{wrapfigure}

We use SONIC's VR whole-body teleoperation interface from GR00T-WholeBodyControl~\citep{sonic2025}, operated in its three-point (\texttt{vr3pt}) tracking mode.
The operator wears a PICO~VR device: the headset and two hand controllers supply head and dual-wrist poses that SONIC retargets to the wrist targets, the controller triggers set the per-arm grip, and the joysticks set the 4-D base command ($v_x, v_y, \omega_z, h$; Figure~\ref{fig:teleop}).
SONIC consumes these inputs and produces the shared 18-D high-level command stream driving both the simulator and the real robot.
The 50 real demonstrations per task were collected over multiple sessions; episodes with fall events or hardware safety triggers were discarded and re-collected to maintain a balanced 50-episode budget, and this re-collection is included in the operator-hour figure of Appendix~\ref{app:dataset_cost}.

\subsection{Primitive-Based Generation Protocol}

\paragraph{Primitive composition.}
Each pick-and-place task decomposes into three high-level motions: \texttt{Walk}, \texttt{Pick}, and \texttt{Place}. \texttt{Walk} takes a target object and runs \texttt{Align}, \texttt{Approach}, then \texttt{Align} again to position the robot in front of the object; the final \texttt{Align} commands base yaw ($\omega_z$) to refine the robot's heading toward the active arm---a pre-fixed grasping arm for \texttt{Pick} or the arm currently holding the object for \texttt{Place}---so the target falls into a better reach angle for that arm. A \texttt{Stepback} primitive runs first when \texttt{Walk} is invoked while holding an object. \texttt{Pick} and \texttt{Place} share a common opening, with \texttt{Stand/Squat} adjusting the pelvis height to the object height $h_\mathrm{obj}$ and \texttt{Reach} interpolating the end-effector toward an initial pose estimate of the target.

\texttt{Pick} then closes the loop with an \texttt{Adapt} primitive that corrects the gripper pose using the commanded-vs-actual wrist offset, a privileged simulator signal used only during data generation and not at deployment; on a closure-tolerance failure the arm retracts to a re-pose and re-runs \texttt{Reach} and \texttt{Adapt}. Each motion ends with a verification primitive---\texttt{VerifyHold} after \texttt{Pick} and \texttt{VerifyPlace} after \texttt{Place}---that confirms success before the episode is recorded.

\paragraph{Execution protocol.}
For each task, the scene config declares a \texttt{task\_plan}, an ordered list of task primitives, and the manager executes the plan against the current scene state. Each primitive reports whether it is still running, has completed, or has failed; on failure the manager applies primitive-specific retries (e.g., re-attempting an \texttt{Approach} with a slightly offset waypoint) before declaring episode failure. Successful episodes are retained toward the 200-demo budget; episodes that fail verification or any primitive are discarded and not counted.

\paragraph{Action encoding and verification parameters.}
The 18-D command is ordered as per-arm blocks of [wrist 6-D pose, grip], left arm then right arm, followed by the base command $(v_x, v_y, \omega_z, h)$. Wrist poses are expressed relative to the pelvis, with rotations as Euler-XYZ angles, and commands are issued at 30~Hz. \texttt{VerifyHold} uses a 0.10~m object--wrist tolerance along $z$; \texttt{VerifyPlace} requires the object to lie within 0.10~m in $xy$ and 0.06~m in $z$ of the target surface. \texttt{Pick} allows two grasp-servo (\texttt{Adapt}) iterations and one outer retry.

\paragraph{Task specification and generation yield.}
\label{app:supervision}
\ours removes teleoperation and per-episode human intervention, but each task is still specified by hand. Shared across all tasks and built once are the motion primitives, verification predicates, module-level tolerances and retry limits, the episode manager, the controller interface, and color calibration. Specified per scene or task are the scene geometry and object roles, the ordered task plan, the grasping arm and initial target poses, the language prompt, the randomization ranges, and per-object grasp parameters. This per-scene and per-task part takes under 15 operator-minutes, in addition to an unattended \gsplat reconstruction. Within the pick-and-place family we reuse the existing primitives and retune grasp parameters per object; a new task family would require new primitives. Building the framework is a one-time cost that is not repeated per task and is not included in Tables~\ref{tab:teleop_vs_legs} and~\ref{tab:dataset_cost}. Across Tasks~1--3, retaining 200 verified episodes took 213--235 generated attempts, a yield of 85--94\%.

\section{Task 3 Trajectory and Per-Condition Egocentric Views}
\label{app:task3}

\begin{figure}[p]
  \centering
  \includegraphics[width=\linewidth]{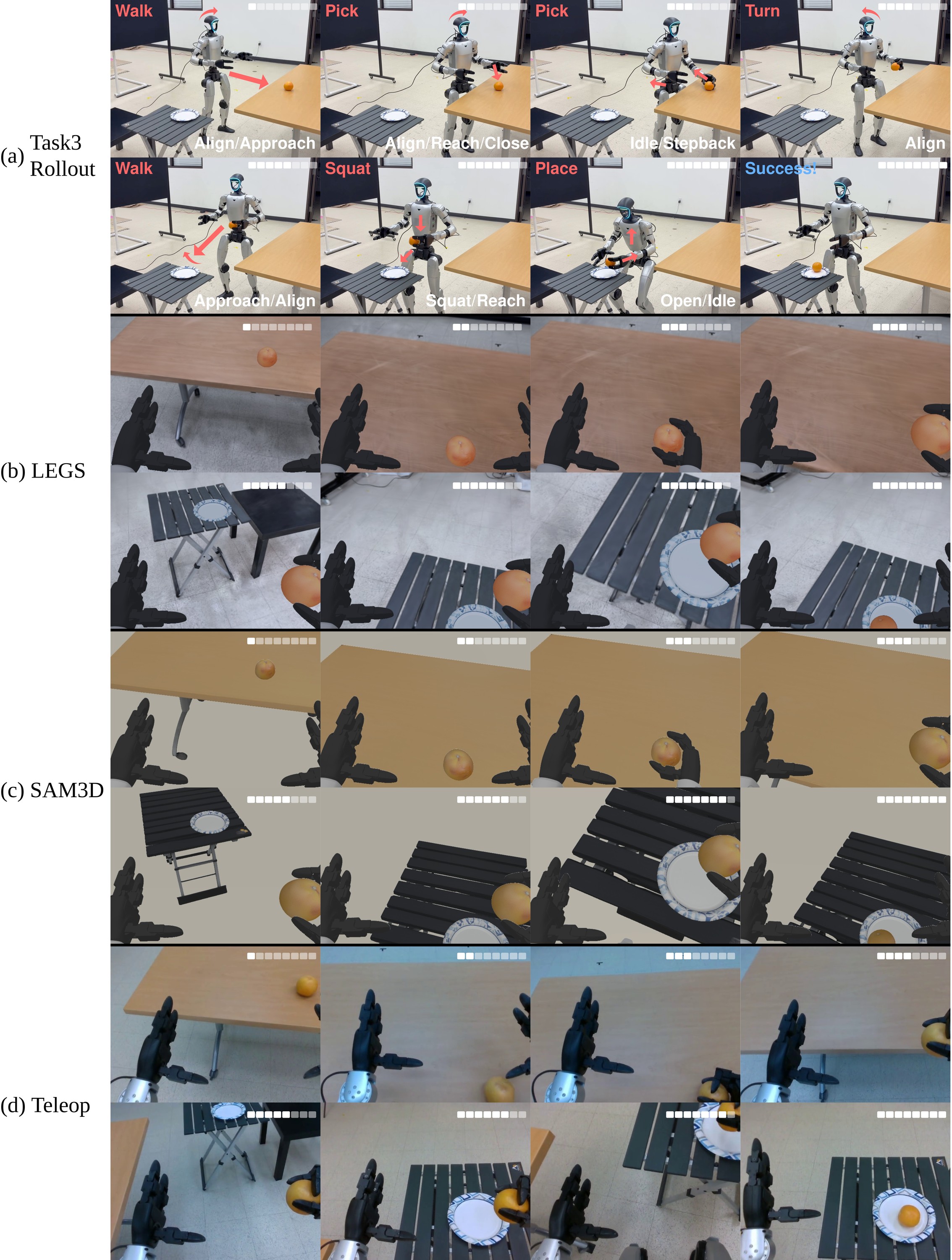}
  \caption{
    \textbf{Task~3 trajectory and per-condition egocentric observations.}
    \textbf{(a)}~Eight third-person keyframes of a successful Task~3 rollout. Each frame is labeled with its \emph{stage} (top-left, red), matching the stages of the stage-wise success plot (Appendix~\ref{app:stagewise}), and the active \emph{motion primitive(s)} (bottom-right, white), which compose the \texttt{Walk}/\texttt{Pick}/\texttt{Place} high-level motions of Appendix~\ref{app:data}.
    \textbf{(b--d)}~Head-camera observations at the same eight keyframes for \ours, the SAM3D mesh-only baseline, and human teleoperation. Panels (b, c) re-render one shared motion and are pose-aligned; (d) is a representative teleoperated rollout. Note (d) is captured with the RealSense auto-exposure/auto-white-balance enabled, whereas \ours~(b) is calibrated to the fixed deployment ISP, so the two are not a like-for-like color match.
  }
  \label{fig:task3_frames}
\end{figure}

Figure~\ref{fig:task3_frames} documents Task~3, our longest-horizon task, at two levels of granularity and across the three data conditions.

\paragraph{Stages and primitives (a).}
The top two rows show eight third-person keyframes from a \groot policy trained on \ours data, each carrying two labels: a coarse \emph{stage} in the top-left (red), which is exactly the stage used in the stage-wise success plot of Appendix~\ref{app:stagewise} (\texttt{Walk}, \texttt{Pick}, \texttt{Turn}, \texttt{Walk}, \texttt{Squat}, \texttt{Place}), and the underlying \emph{motion primitive(s)} in the bottom-right (white), which realize those stages and compose the \texttt{Walk}/\texttt{Pick}/\texttt{Place} high-level motions of Appendix~\ref{app:data}. Read together they trace the full task: \texttt{Walk} to the orange (\texttt{Align}/\texttt{Approach}) and \texttt{Pick} it (\texttt{Reach}/\texttt{Close}, then \texttt{Idle}/\texttt{Stepback}); \texttt{Turn} toward the low table (\texttt{Align}); a second \texttt{Walk} across to it (\texttt{Approach}/\texttt{Align}); and \texttt{Squat}/\texttt{Place} to lower the pelvis and release the orange (\texttt{Squat}/\texttt{Reach}, then \texttt{Open}).

\paragraph{Per-condition egocentric views (b--d).}
The lower rows show the egocentric views at the same eight stages. Panels (b) and (c) re-render the \emph{same} recorded motion, so they are pose-aligned frame-by-frame and differ only in rendering frontend---\ours composites mesh foreground over the \gsplat background with color calibration, while SAM3D renders the full scene (Appendix~\ref{app:sam3d}); panel (d) is a representative teleoperated rollout. The contrast between (b) and (c) illustrates the close-range gap of Section~\ref{sec:photorealism}: \ours's \gsplat background renders the room photorealistically, whereas the SAM3D render is comparatively flat and uniform. Panel (d) is shown only as a reference for the real task; because teleop runs the RealSense with auto-exposure and auto-white-balance enabled while \ours is deployed at a fixed ISP, (b) and (d) differ in exposure and color and are not directly comparable pixel-for-pixel.

\section{Matched-Camera Evaluation on Task 3}
\label{app:matched}

In Table~\ref{tab:main_tsr}, \ours policies are deployed with fixed exposure and white balance, while Teleop and SAM3D policies use the RealSense's automatic mode. To remove this confound, we re-evaluate the same \ours checkpoints with the camera in automatic mode, which also lies outside the regime in which the color-correction matrix was fitted. We use the long-horizon Task~3 and extend every cell to $n{=}20$ trials; for the conditions in Table~\ref{tab:main_tsr}, the first 10 trials are those reported there. Initial conditions are drawn independently per trial from the ranges of Appendix~\ref{app:randomization}.

\begin{table}[h]
\centering
\small
\setlength{\tabcolsep}{5pt}
\caption{\textbf{Task~3 end-task success under matched and fixed camera settings, $n{=}20$ per cell.} ``Auto'' is the RealSense automatic exposure and white-balance mode used by the baselines; ``fixed'' is the \ours deployment setting.}
\label{tab:matched}
\begin{tabular}{l ccc c}
\toprule
Data / deployment ISP & \psizero & \pihalfpoint & \groot & All \\
\midrule
Teleop (50) / auto   & 1/20 & 0/20 & 0/20 & 1/60 \\
SAM3D (200) / auto   & 2/20 & 0/20 & 4/20 & 6/60 \\
\ours (50) / auto    & 5/20 & 1/20 & 8/20 & 14/60 \\
\ours (200) / auto   & 8/20 & 1/20 & 9/20 & 18/60 \\
\ours (50) / fixed   & 5/20 & 2/20 & 10/20 & 17/60 \\
\ours (200) / fixed  & \textbf{9/20} & \textbf{3/20} & \textbf{11/20} & \textbf{23/60} \\
\bottomrule
\end{tabular}
\end{table}

Under matched automatic settings, \ours still leads at both data budgets. With Cochran--Mantel--Haenszel tests stratified by backbone, \ours~(50) exceeds Teleop~(50) ($p{=}0.0008$) and \ours~(200) exceeds SAM3D~(200) ($p{=}0.009$). Holding the checkpoint fixed, moving \ours~(200) from fixed to automatic camera settings reduces success from 23/60 to 18/60, a difference that is not statistically distinguishable under the same stratified test ($p{=}0.41$). On \pihalfpoint every automatic-mode condition is at or below 1/20, a floor effect. Because \ours and SAM3D still differ in both background representation and color calibration, this comparison establishes a frontend-level effect only.

\section{Policy Training Details}
\label{app:training}
All three backbones share SONIC as the whole-body controller (Section~\ref{sec:physics_backend}) and emit the same 18-dimensional command, so within-backbone comparisons isolate the data source. Each checkpoint is trained with a single seed, and \ours~(50) is a fixed first-50 prefix of \ours~(200), so we do not estimate seed or subset variance. They differ only in fine-tuning scope and the per-backbone hyperparameters of Table~\ref{tab:training}.

\begin{table}[h]
\centering
\small
\caption{
  \textbf{Per-backbone fine-tuning configuration.}
  All runs: bf16 mixed precision, 20{,}000 optimization steps, 4$\times$H100~80GB GPUs, 18-D (EEF + base) action, peak learning rate $10^{-4}$ with a 1k-warmup cosine schedule, and flow-matching loss.
}
\label{tab:training}
\begin{tabular}{l p{2.7cm} p{2.7cm} p{3.8cm}}
\toprule
 & \psizero & \pihalfpoint & \groot \\
\midrule
Fine-tuning scope   & action head only        & full (VLM + head)      & action head only \\
Action chunk        & 30                      & 16                     & 16 \\
Effective batch     & 256                     & 128                    & 24 \\
Checkpoint cadence  & every 2{,}500 steps     & every 2{,}500 steps    & every 10{,}000 steps \\
Initialization      & \psizero & DROID-pretrained $\pi_{0.5}$ & GR00T-N1.6-3B \\
\bottomrule
\end{tabular}
\end{table}

\end{document}